\documentclass[10pt,twocolumn,letterpaper]{article}
\pdfoutput=1
\makeatletter
\renewcommand*{\@fnsymbol}[1]{\ensuremath{\ifcase#1\or \dagger\or \ddagger\or
		\mathsection\or \mathparagraph\or \|\or **\or \dagger\dagger
		\or \ddagger\ddagger \else\@ctrerr\fi}}
\makeatother
\usepackage{cvpr}
\usepackage{times}
\usepackage{epsfig}
\usepackage{graphicx}
\usepackage{amsmath}
\usepackage{amssymb}
\usepackage{url}
\usepackage{pifont}
\usepackage{multirow}
\usepackage{bm}
\usepackage{breqn}
\usepackage[breaklinks=true,bookmarks=false]{hyperref}

\cvprfinalcopy 

\def\cvprPaperID{****} 

\begin{document}
	
	\title{Self Attention Grid for Person Re-Identification}
	
	\author{\quad Jean-Paul Ainam \thanks{Jean-Paul Ainam is also a lecturer at Cosendai University, Cameroon. He is currently a Ph.D. student at the University of Electronic Science and Technology of China.} 
		\qquad \quad \qquad \qquad  Ke Qin\thanks{Co-corresponding authors.} 
		\qquad \quad \qquad \quad Guisong Liu\footnotemark[2]
		\\
		{\tt\small jpainam@uacosendai-edu.net} 
		\qquad {\tt\small qinke@uestc.edu.cn}
		\qquad {\tt\small lgs@uestc.edu.cn}
		\\ 
		School of Computer Science and Engineering \\
		University of Electronic Science and Technology of China \\
		Chengdu, Sichuan, P.R. China, 611731
	}
	
	\maketitle
	
	\begin{abstract}
		In this paper, we present an attention mechanism scheme to improve person re-identification task. Inspired by biology, we propose Self Attention Grid (SAG) to discover the most informative parts from a high-resolution image using its internal representation. In particular, given an input image, the proposed model is fed with two copies of the same image and consists of two branches. The upper branch processes the high-resolution image and learns high dimensional feature representation while the lower branch processes the low-resolution image and learn a filtering attention grid. We apply a max filter operation to non-overlapping sub-regions on the high feature representation before element-wise multiplied with the output of the second branch.  The feature maps of the second branch are subsequently weighted to reflect the importance of each patch of the grid using a softmax operation. Our attention module helps the network learn the most discriminative visual features of multiple image regions and is specifically optimized to attend feature representation at different levels. Extensive experiments on three large-scale datasets show that our self-attention mechanism significantly improves the baseline model and outperforms various state-of-art models by a large margin. 
		
	\end{abstract}
	
	\section{Introduction}
\begin{figure}[t]
		\centering
		\includegraphics[width=\linewidth]{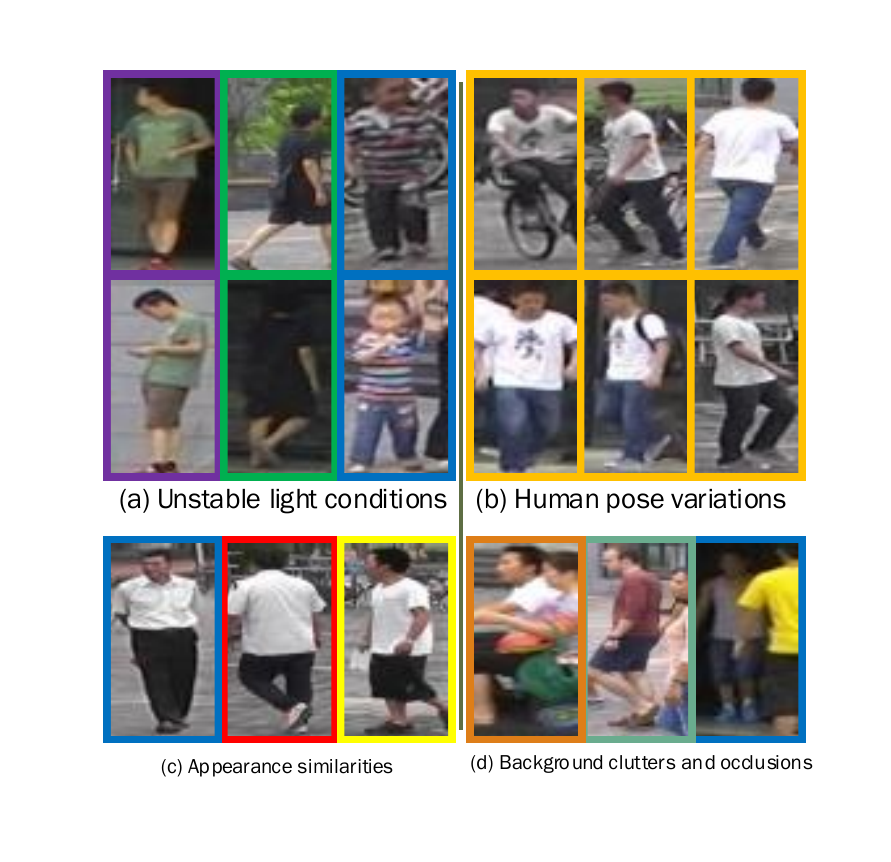}
		\caption{Person re-id challenges. Images with a same border color represent the same identities within a given group.}
		\label{fig:challegences}
	\end{figure}

	Person re-identification is the problem of identifying persons across images using different cameras or across time using a single camera. Automatic person re-identification has become essential in surveillance systems due to the rapid expansion of large-scale distributed multi-camera systems. However, many issues still prevent person re-id of achieving high accuracy as compared to other image recognition tasks; its performance is still far from optimal. These issues relate to the fact that person re-id usually needs to match the person images captured by surveillance cameras working in wide angle mode with a very low resolution and unstable lighting conditions. Despite the increasing attention given by researchers to solve the person re-id problem, it remains a challenging task in a practical environment. Some of these challenges are depicted in \ref{fig:challegences} and are: dramatic variations in visual appearance and ambient environment (a), human pose variation across time and space (b), background clutter and occlusions (c), and different individuals sharing similar appearances (d) among others.
	
	Current approaches to solving person re-id generally follow a verification or identification framework or both \cite{Varior2016Gated, zheng2016discriminatively}. Such a framework takes as input a pair of images and outputs a similarity score or a classification result. Moreover, a Siamese convolution neural network architecture \cite{Li2014Pairing, Varior2016Gated,Zhao2017PartAligned, zheng2016discriminatively} which consists of two copies of the same network has recently emerged. The two networks are connected by a cost function used to evaluate the relationship between the pair. Other architecture models, driven by a triplet loss function, have resulted into part-based networks \cite{Cheng2016, Li2017DeepContext, Sun2017RPP} where the first convolution layers learn low-level features while fully connected layers concentrate on learning higher-level features. All the part contribute to the training process jointly.  
	In the last decade, with the advent of deep generative models, GAN-based models \cite{yu2017cross, Zhang2018Crossing, zheng2017unlabeled, zhong2018camera} have slightly increased the performance of person re-id task; however, results from these works show that there is still a room for improvement. 
	
	In this paper, inspired by biology and the recent success of attention mechanism on Recurrent Neural Network \cite{Chorowski2015,Hermann2015,Mnih2014HardAttention,Shan2017Speech, Zeyer2018Speech} and Convolution Neural Network \cite{Jaderberg2015STN, Li2018Harmonious, Liu2017End2End, Rahimpour2017Attention,Rodriguez2017AgeGender, Wang2017ResAttention, Wu2018CoAttention, Xu2016AskAttendAnswer, Yang2016StackedAttention, Zhang2016Picking, Zhao2017Diversified}, we propose  Self Attention Grid (SAG) for person re-identification. Attention mechanism gives a network a capability to focus on specific parts of the visual input to compute the adequate response. In other words, It helps the network select the most pertinent piece of information, rather than using all available information. This is particularly important in person re-id where input images might contain information that is actually irrelevant for computing visual responses. In a classical person re-id model, the whole input image is used to predict the output regardless of the fact that not all pixels are equally important. As a result, we use a variant of self-attention mechanism to overcome such limitations. The contributions of this work are:
	\begin{enumerate}
		\item a simple feed-forward attention mechanism with multiple non-overlapping attention regions;
		\item an attention module solely based on self-attention that can extract high discriminative feature representation from a high-resolution image and preserve low-level information from the initial representation;
		\item a fully differentiable attention module that can be easily plugged into any existing network architecture without much effort.
	\end{enumerate}
	\section{Related Works}
	In this section, we present the works relevant to understand our approach. This include attention mechanism background together with their application to person re-identification task.  
	
	\subsection{Attention Mechanism}
	Inspired by neuroscience and biology, attention mechanism allows deep neural network to focus on specific parts of the input vector. Not only it allows a network to learn the most discriminative feature, but also it effectively reduces the computational burden of processing the whole input vector.
	
	Attention mechanism started in the field of Natural Language Processing (NPL)  \cite{Bahdanau2015} and became central to many deep learning approaches, especially Recurrent Neural Networks (RNNs). It has been successfully applied to various interesting tasks such as text-based question answering \cite{Hermann2015}, image captioning \cite{Xu2015ShowAttendTell}, visual question answering \cite{Yang2016StackedAttention, Xu2016AskAttendAnswer}, speech recognition \cite{Chorowski2015, Shan2017Speech, Zeyer2018Speech} and fine-grained image classification \cite{Jaderberg2015STN, Wang2017ResAttention, Zhang2016Picking, Zhao2017Diversified}. Following its success in machine translation, many researchers started exploring its application into the computer vision fields by proposing various forms of attention mechanisms: hard-attention \cite{Mnih2014HardAttention}, soft-attention, global attention and local-attention to cite but a few. 
	
	In particular, self-attention mechanism, also referred to as \textit{intra-attention} in \cite{Cheng2016LSTMMR, Parikh2016Decomposable} attends to different parts of a single sequence by using the internal representations of the same sequence. \cite{Vaswani2017Attention} proposed scaled dot-product attention combining self-attention with a scaling factor and successfully achieved state-of-art in machine translation. \cite{Parmar2018Transformer} generalized an autoregressive model architecture based on self-attention for image generation and \cite{Xiaolong2018NonLocal} formalized self-attention for machine translation as a class of non-local filtering operation that can be applied to video sequences.
	
	While many researchers \cite{Liu2017End2End, Li2018Harmonious, Rahimpour2017Attention, Wu2018CoAttention} have investigated the application of soft and hard attention mechanism to person re-identification; however, to the best of our knowledge, multi-depth regions solely based on self-attention mechanism has not yet been explored in this context. Our work, exploit the attention provided by the attention mechanism at different levels to capture information and efficiently learn to focus on specific part of the image by using only the internal representations of the same image at each time.
	
	\subsection{Person Re-Identification} 
	Person re-id works can be roughly divided into two groups: distance metric learning and deep machine learning based approaches. The first group, also named discriminative distance metric focus on learning local and global feature similarities by leveraging inter-personal and intra-personal distances \cite{ Chen2016, Kostinger2012, Liao2015, Liao2015a, Xiong2014, Zhang2011, Zheng2015}. The second group is CNN-based with a goal to jointly learn the best feature representation and a distance metric. Some feature learning approaches \cite{Cheng2016, Li2017DeepContext, Sun2017RPP} decompose the images into part based. Other methods \cite{Li2014Pairing, Varior2016Gated, Zhao2017PartAligned, zheng2016discriminatively} used a siamese convolution neural network architecture for simultaneously learning a discriminative feature and a similarity metric. Given a pair of input images, they predict if it belongs to the same subject or not through a similarity score. To improve the similarity score, \cite{Paisitkriangkrai2015, Zhong2017reranking} proposed to optimize the evaluation metrics commonly used in person re-id. Recently \cite{yu2017cross, Zhang2018Crossing, zheng2017unlabeled, zhong2018camera} proposed to address the problem of lack of large datasets in person re-id by training a CNN based architecture and a GAN \cite{Goodfellow2014GAN} generated samples through a regularization method \cite{Szegedy2016}. It was particularly observed that generated images improve the re-id accuracy when combined with a training sample.
	
	Similar to our work, \cite{Liu2017End2End} proposed an end-to-end Comparative Attention Network (CAN) to progressively compare the appearance of a pair of images and determine whether the pair belongs to the same person. During training, a triplet of raw images is fed into CAN for discriminative feature learning and local comparative visual attention generation. Their network architecture, made of two parts require much higher computation cost compared to our work. \cite{Li2018Harmonious} proposed a complex CNN model for jointly learning soft and hard attention. The two attention mechanisms with feature representation learning are simultaneously optimized. Finally, \cite{Rahimpour2017Attention} proposed gradient-based attention mechanism to solve the problem of pose and illumination found in re-id problem in a triplet architecture and \cite{Wu2018CoAttention}  recommended Co-attention based comparator to learn a co-dependent feature of an image pair by attending to distinct regions relative to each pair. 
	
	In general, attention mechanism is used in re-identification task to discover the most discriminative information for further processing. 
	We depart from these works and propose our attention based-CNN model in next section.
	\section{Our Approach}
	\begin{figure*}
		\centering
		\includegraphics[width=\linewidth]{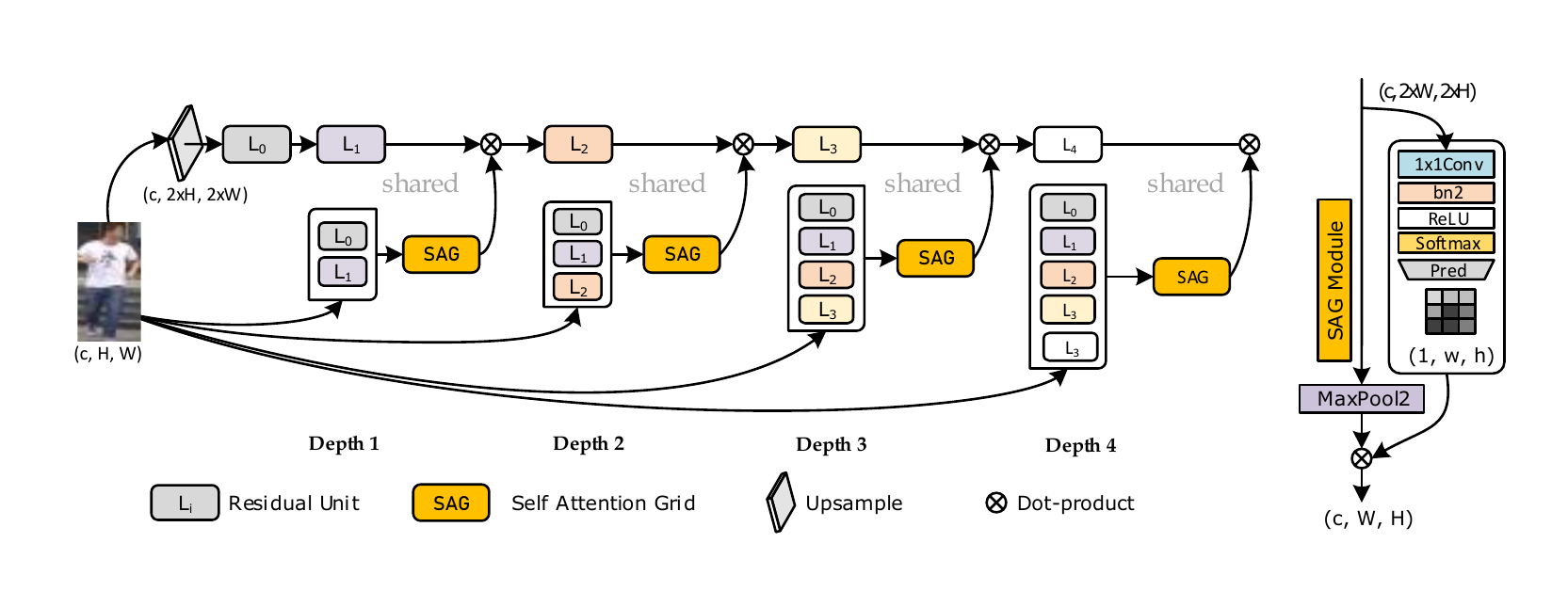}
		\caption{The baseline architecture model (left) coupled with our Self Attention Grid (SAG) Module (right). We show how SAG modules can be easily integrated into existing network without much efforts.}
		\label{fig:framework_module}
	\end{figure*}
	In this section, we introduce our Self Attention Grid (SAG) mechanism coupled with ResNet50 baseline  specifically designed for person re-identification tasks. We first describe the overall network architecture and then elaborate on the design of the SAG module.
	\subsection{Overview}
	As described in Figure \ref{fig:framework_module}, the overall network architecture of the proposed attention consists of two branches sharing the same weights. After each residual unit (\textit{layer$_i$}), we introduce a SAG module, we refer to this position as the attention depth ($D_i$). The network is fed with two copies of the same image $I_1 \in \mathbb{R}^{C\times H_1 \times W_1}$ and $I_2 \in \mathbb{R}^{C \times H_2 \times W_2}$ such that the spatial dimension of $I_1$ is twice the spatial dimension of $I_2$ i.e $(H_1,W_1)=(2H_2, 2W_2)$. To do this, we upsampled $I_1$ by a factor of two using bilinear interpolation as follows:
	\begin{equation} \label{eq:bilinear}
	\begin{split}
	H_{1} & = \left \lfloor H_{2} \times \text{scale\_factor} \right\rfloor \\
	W_{1} & = \left \lfloor W_{2} \times \text{scale\_factor} \right\rfloor
	\end{split}
	\end{equation} 
	where $\text{scale\_factor}=2$, $(H_1,W_1)$ the height and width of the high resolution  image $I_1$, $(H_2, W_2)$ the spatial dimension of the original image $I_2$ and $C$ the channel.
	
	At depth $D_i$, the first branch (upper-branch) processes the high resolution image $I_1$ and outputs a high dimensional features $f_{1}^{D_i}$, whereas the second branch (lower-branch) processes the low-resolution image $I_2$ and outputs a low dimensional features $f_{2}^{D_i}$ representing the filtered feature map, with our filtering attention grid putting focus on the interesting part of the original image. Given $D_i$, the network computes the attention response $f_{2}^{D_i}$ weighted by an importance score predicted by the SAG modules. In the next sections, for the sake of brevity we simply refers to $f_{1}^{D_i},f^{D_i}_{2}$ as $f_1, f_2$. 
	
	Given the high dimension image, we learn a discriminative feature using the upper CNN branch before element-wise multiplied by the output of the attention grid coming from the lower branch. The proposed method goes beyond the traditional CNN based attention models in re-identification and proposes an attention grid network that can learn multiple discriminative parts of person images as depicted in Figure \ref{fig:multiregions}.
	
	In general, attention model output a summary vector $z$ of a class probability $y_i$; focusing on the information of an input vector $x_i$. $z$ is usually a weighted arithmetic mean of $y_i$; with weights chosen by relevance of each $y_i$, given $x_i$. In our case, the output of the attention module is a multi-grid region which relevance is produced by a softmax operation.
	
	The overall network architecture is finally trained at once for identity classification task using supervised learning. We use the conventional cross-entropy loss function defined by:
	\begin{equation} \label{eq:crossentropy}
	\mathcal{L}(\theta) = - \sum_{i=1}^{N} \log p(\tilde{y}_i=y_i)
	\end{equation}
	where $N$ denotes the number of output classes, $p(\tilde{y}_i=y_i)$ the vector class probability produced by the neural network, $\tilde{y}_i$ the predicted label, $y_i$ the ground-truth label and $\theta$ the network parameters. The network is trained to minimize Equation \ref{eq:crossentropy}
	\subsection{The Self Attention Grid} \label{selfattentiongrid}
	Let $X=\{(x^{(1)}, y^{(1)}),(x^{(2)}, y^{(2)}), \cdots (x^{(n)}, y^{(n)})\}$, be $n$ training samples where $x^{(i)}$ represents an image $\in \mathbb{R}^{C\times H \times W}$ and $y^{(i)} \in [1, \ldots, n]$ its corresponding label; we aim at learning a feature representation model for person matching across multiple views. As a result, we propose Self Attention Grid as an attention mechanism for locating the most discriminative pixels and regions at different depths. We consider a multi-branch network architecture with weights shared between the two branches.
	
	Our Self Attention Grid module consists of a $1\times1$ convolution, a batch normalization, an activation function (ReLU) and a softmax function. A convention neural network represents a function $h(x;\theta)$ defined by:
	\begin{equation}
	h(x) = \bm{w}^T_k\cdot x + b_k
	\end{equation}
	where $(\bm{w}_k, b_k)$ denote the network parameters that are automatically learned from data. Given $x_i$, the output $h(x_i)$ of a residual unit $L_i$ is fed into the $1\times1$ convolution layer to produce an attention heatmap $e_{ij}$. We use the softmax operation to normalize the full self attention so as to make the sum equal 1 using:
	\begin{equation}
	S_{ij} = softmax(e_{ij}) = \frac{\exp(e_{ij})}{\sum_{j=0}^{N} \exp(e_{ij})}
	\end{equation}
	The softmax layer computes the maximum of relevance of the variable $x^{(i)}$ according to $e{ij}$. 
	$S_{ij}$ is a two dimensional vector which we then convert into a grid attention $\bm{G}_{ij}$ $\in \mathbb{R}^{N\times 1\times h \times w}$ such that :
	\begin{equation}
	\sum_{i=1}^{N} \bm{G}_{ij} = 1
	\end{equation}
	The output of the each attention grid consists of one spatial channel dimension and defines a function $\phi : \mathbb{R}^{N \times C \times H \times W} \rightarrow \mathbb{R}^{N \times 1 \times h \times w}$. The output of the attention module $f_2$ is element-wise multiplied by each channel dimension of $f_1$ produced by the same layer sequence processing the high image resolution:
	\begin{equation} \label{eq:sagproduct}
	v = f_1 \odot f_2
	\end{equation}
	where $f_1 \in \mathbb{R}^{C \times H \times W}$ and $f_2 \in \mathbb{R}^{1 \times h \times w}$. To make this possible, we applied a sampled-based discretization process to downsample the high feature representation $f_1$, thus reducing its dimensionality and allowing the feature contained in the sub-regions of $f_1$ to correspond to the initial sub-region representations of $f_2$.
	
	Finally, we performed $L_2$ normalization over the attended regions $v$ using:
	\begin{equation} \label{eq:normalization}
	f^I =  \frac{v}{\max(\lVert v\rVert_2, \epsilon)}
	\end{equation}
	where $\lVert \cdot \rVert_2$ is the Euclidean norm and $\epsilon = 1e-12$ a small value to avoid division by zero. This Equation \ref{eq:normalization} helps keep the overall error small. We show in section \ref{discussion}, how this can be used to improve the classification accuracy.
	
	In general, our approach can be considered as a kind of region-based-attention for it searches through the image multiples regions that match what the network is interested in for further processing.
	
	\begin{table*}[t]
		\centering
		\caption{Impact depth of SAG modules on three datasets. We report results with \textbf{no} re-ranking \cite{Zhong2017reranking}. Basel.+ $D_i$ refers to our baseline architecture with SAG module at depth \textit{i}}
		\label{tab:ablationresults}
		\resizebox{\textwidth}{!}{%
			\begin{tabular}{|l|c|c|c|c|c|c|c|c|c|c|c|c|}
				\hline
				\multirow{2}{*}{SAG Components}& \multicolumn{4}{c|}{Market-1501 \cite{Zheng2015}} &  \multicolumn{4}{|c|}{CUHK03 \cite{Li2014Pairing}} & \multicolumn{4}{c|}{DukeMTMC \cite{zheng2017unlabeled}}\\
				\cline{2-13}
				&R1&R5&R10&mAP&R1&R5&R10&mAP&R1&R5&R10&mAP\\
				\hline
				Baseline(ResNet50)&83.49&93.82&95.90&63.87&71.04&92.89&96.63&80.46&72.04&84.96&89.09&54.36\\
				Basel.+ $D_1$ (h=40,w=16)&84.35&94.00&96.08&64.40&78.50&95.60&98.22&85.93&73.74&84.87&88.24&51.99\\
				Basel.+ $D_2$ (h=20,w=8)&85.18&94.12&96.20&65.90&78.99&95.63&98.13&86.25&74.28&85.05&89.18&52.20\\
				Basel.+ $D_3$ (h=10,w=4)&87.65&95.07&96.91&70.03&81.41&96.62&98.51&88.00&76.30&87.25&90.80&56.62\\
				Basel.+ $D_4$ (h=5,w=2)&\textbf{90.17}&\textbf{96.38}&\textbf{97.48}&\textbf{73.87}&\textbf{82.46}&\textbf{96.44}&\textbf{98.42}&\textbf{88.64}&\textbf{79.94}&\textbf{89.68}&\textbf{92.15}&\textbf{60.88}\\
				Basel.+ $D_{1,2}$&84.09&93.97&96.17&61.28&77.60&95.79&98.50&85.52&74.55&84.96&88.69&51.65\\
				Basel.+ $D_{1,2,3}$&85.10&94.24&96.23&65.60&77.77&95.81&98.34&85.59&72.44&84.52&88.47&51.03\\
				Basel.+ $D_{1,2,3,4}$&82.13&92.79&95.72&60.81&76.51&94.86&97.86&84.53&70.42&82.27&86.58&4876\\
				\hline
			\end{tabular}
		}
	\end{table*}
	\section{Experiments} \label{experiments}
	To validate the effectiveness of the proposed attention mechanism, we intensively conduct experiments and ablation study on three widely used datasets \footnote{The code of the experiments is available at \url{https://github.com/jpainam/self_attention_grid}}. 
	
	\subsection{Person Re-ID Datasets}
	Three datasets were used to evaluate the proposed attention scheme. They include Market-1501, CUHK03 and DukeMTMC-ReID.
	
	\textbf{Market-1501} \cite{Zheng2015} is one of the largest and most realistic dataset in person re-identification. It is collected using six overlapping cameras. The image bounding boxes were automatically detected using the Deformable Part Model (DPM) \cite{Felzenszwalb2010}. The dataset contains $32,668$ from $1,501$ identities divided into $12,936$ images for the training set and $ 19,732$ images for the testing. There are $751$ identities in the training set, $750$ identities in test set, $3,368$ query images and $2,793$ distractors. In this work, we use all the training set for training and all the test set for testing. 
	
	\textbf{CUHK03} \cite{Li2014Pairing} contains $13,164$ images from $1,467$ identities. The dataset is captured by six cameras, but each identity only appears in two disjoint camera views with an average of $4.8$ images in each view. The dataset is split into two subsets, one set contains manually cropped bounding boxes, and the other set is automatically detected using the Deformable Part Model \cite{Felzenszwalb2010}. In this work, we use the detected set.
	
	\textbf{DukeMTMC-reID} \cite{zheng2017unlabeled} is a subset of a pedestrian tracking dataset DukeMTMC \cite{Ristani2016}. The original dataset is a collection of handcrafted bounding boxes and high resolution videos data set recorded by 8 synchronized cameras over $2,000$ identities. In this work, we use the subset defined in \cite{zheng2017unlabeled}. The subset follows the Market-1501 format and contains $36,411$ images from $1,404$ identities divided into $16,522$ images from $702$ identities for the training set and $17,661$ images from $702$ identities for the test set. There are $2,228$ query images and $17,661$ gallery images.
	
	For a fair comparison, we follow the dataset split strategy of each dataset as described in their first released. Table \ref{tab:datasets} gives a summary of the split strategy.
	\begin{table}
		\centering
		\caption{Dataset split details. The total number of images (\textit{QueryImgs, GalleryImgs, TrainImgs}), together with the total number of identities (\textit{TrainID, TestID}) are listed.}
		\label{tab:datasets}
		\begin{tabular}{|l|c|c|c|c|}
			\hline
			Dataset &Market&CUHK03&Duke\\
			\hline\hline
			Number of IDs&1501&1,467&1404\\
			Number of Images&36,036&14,097&36,411\\
			Cameras&6&2&8\\
			\#Train IDs&751&1367&702\\
			\#Train Images&12,936&13,113&16,522\\
			\#Test IDs&750&100&702\\
			\#Query Images&3,368&984&2,228\\
			\#Gallery Images&19,732&984&17,661\\
			\hline
		\end{tabular}
	\end{table}

	\subsection{Evaluation Metrics}
	We adopt the quantitative metrics cumulative matching curve (CMC) and Mean Average Precision (mAP) as they are commonly used in person re-id.
	
	\textbf{Cumulative Matching Curve} is a precision curve that provides recognition for each rank. Rank-\textit{k} accuracy denotes the probabilities of one or more correctly matched images appearing in top-\textit{k}. Given an query set $Q = \{I_i\}^n_{i=0}$ from $n$-identity, we compute the $L_2$ distance between the query image and all gallery images and return a list of the top-$n$ images. If the returned list contains the query image at a position $k$-th, we consider this query as success at rank-\textit{k} and set it to $1$; if the top-\textit{k} ranked gallery samples do not contain the query identity, we set it to $0$. The final CMC curve is computed by averaging rank-\textit{k} over all the queries.
	
	\textbf{Re-ranking} Recent works \cite{Bai2017,Wang2018, Zhong2017reranking} choose to perform an additional re-ranking to improve the re-identification accuracy. In this work, we use re-ranking with \textit{k}-reciprocal encoding \cite{Zhong2017reranking}, which combines the Euclidian distance and Jaccard distance.
	
	Note that, all the CMC score for the CUHK03 and DukeMTMC datasets are computed with the single-shot setting. Only, experiments on Market-1501 dataset are under both the single-query and multi-query evaluation settings.
	
	\subsection{Implementation details}
	We use ResNet50  \cite{Kaiming2015}, pre-trained on imageNet as baseline and fine-tune the model according to the number of classes i.e. $751; 1,367$ and $702$ units for Market-1501, CUHK03 and DukeMTMC-ReID respectively.  All the input images are resized to $160\times 64$ before random horizontal flipping. We scale the pixels in the range of $-1$ and $1$ and apply zero-center by mean pixel and random erasing \cite{Zhong2017}. Finally, we train the model for $200$ epochs using stochastic gradient descent (SGD) with a batch size of $32$, a momentum of $0.9$ and a weight decay of  $5 \times 10 ^{-4}$. We use a base learning rate $lr$ of $0.01$ for upper layers and $0.1$ for fully connected layers. To further improve the training capability, we gradually decrease $lr$ by a factor of $0.1$ every $30$ epochs using an exponential policy: $lr=lr^{(0)} \times \gamma^{\frac{k}{step\_size}}$ where $lr^{(0)}$ is base learning rate, $\gamma=0.1$, $step size=30$ and $k$ the index of the current mini-batch iteration. We use a validation set to evaluate intermediate models and select the one with maximum performance for testing.
	
	\subsection{Ablation study} \label{ablation}
\begin{figure}
		\centering
		\includegraphics[width=\linewidth]{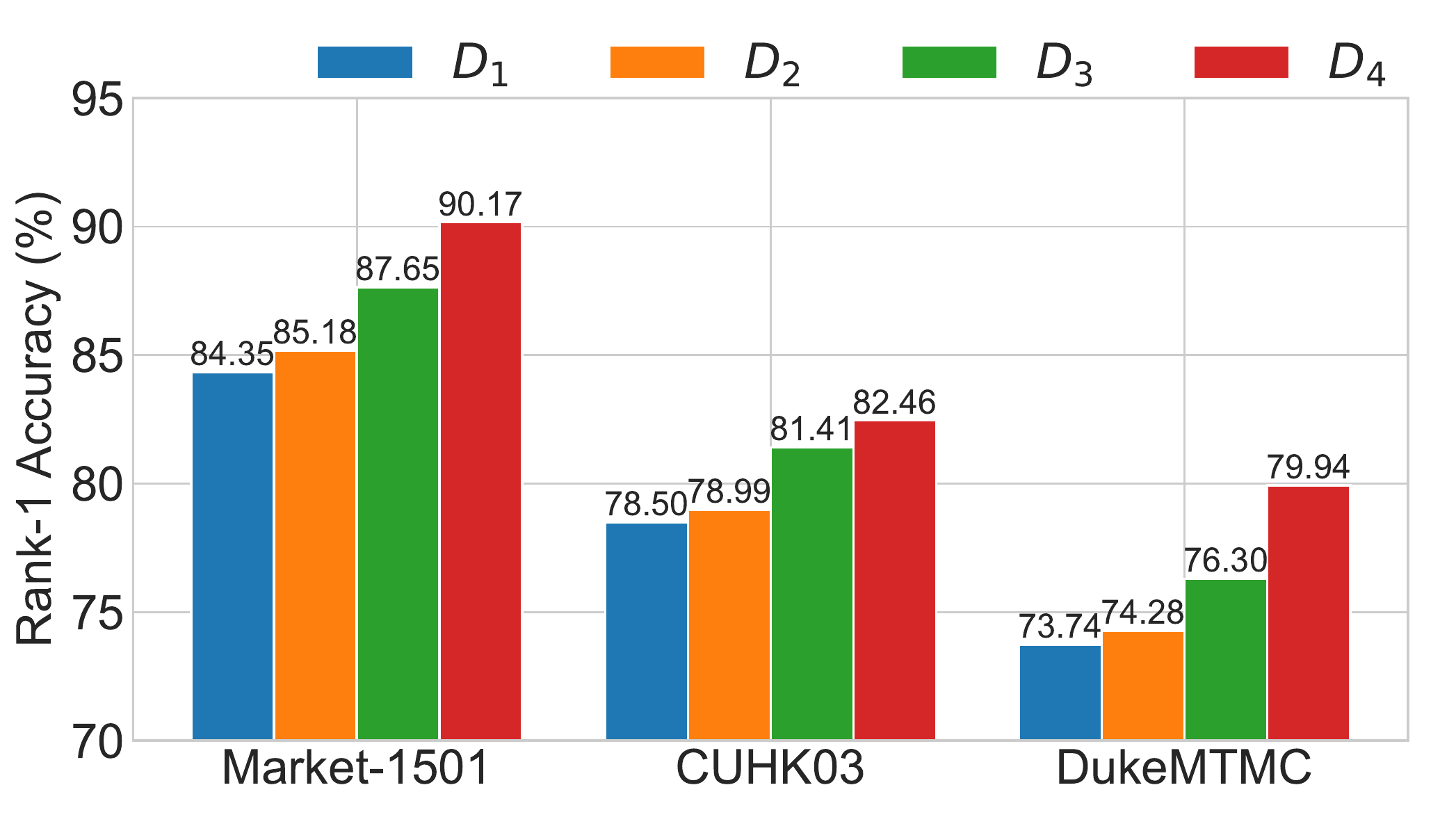}
		\caption{The effect of the position of SAG within a CNN architecture. It can be observed that deeper position yields better results compared to shallow}
		\label{fig:rank1bar}
	\end{figure}

	In this section, we give detailed analysis and investigate the impact produced by SAG module when we vary the attention depth $D_i (i \in [1, \ldots 4])$. We systematically introduce SAG modules at different levels (i.e. after each residual unit) to capture information  and observe accuracy variation. Figure \ref{fig:rank1bar} shows that SAG helps minimize the representation learning risk and improve the descriptive power of the baseline. We first show the effect of applying a single SAG module at different depths and then examine the impact of multiple (up to $4$) SAG modules. We conduct a series of experiments on three datasets and report results in Table \ref{tab:ablationresults} 
	
	\textbf{Baseline model without attention module} In Table \ref{tab:ablationresults}, \textit{Baseline(ResNet50)} refers to the ResNet50 baseline trained for person re-id task. We changed the last fully connected layer to match the number of classes of the three dataset and trained the network using supervised learning. As shown in section \ref{comparison}, the average precision accuracy of our baseline model already outperform previous state-of-art methods.  However, this performance is low compared to our attention based model.
	
	\textbf{Baseline model with single SAG module} The $D_1-D_3$ architectures consist only of three or two convolution layers and are computationally efficient, whereas $D_4$ is deeper ($4$-conv layers), computationally more expensive, but has better performance. As it can be seen in Table \ref{tab:ablationresults}, deep $D_i$ results in better accuracy, reaching saturation at depth $D_4$. It results that, applying our attention module on deeper layers yields better results than on upper layers. These experiments show that SAG enforces the network to learn more discriminative representation. We therefore recommend the application of our module on last layers.
	
	\textbf{Baseline model with an accumulation of SAG modules} We test the importance of multiple attention modules at different depths at the same time by increasingly stacking attention layers at $D_{1-4}$ and after each residual unit. Table \ref{tab:ablationresults} shows that stacking multiple SAG modules at different depth results in a accuracy drop of $1\%$ each time. 

	We achieved the best results with $D_4$ settings on all the datasets.
	\subsection{Comparison with the state-of-arts} \label{comparison}
	Tables \ref{tab:marketresults} \ref{tab:cuhk03results} \ref{tab:dukeresults} show comparison results with state-of-art methods. '-' means that no reported results is available and * means paper on ArXiv but not published.  In the results, \textbf{SAG} represents our method with ResNet50 as baseline and \textbf{SAG+RR} represents our model with re-ranking.
	\begin{table}
		\centering
		\caption{Comparison results on Market-1501.}
		\label{tab:marketresults}
		\begin{tabular}{|l|c|c|c|c|}
			\hline
			Query type& \multicolumn{2}{c|}{Single Query} & \multicolumn{2}{c|}{Multi Query}\\
			\hline
			Methods(\%)&R1&mAP&R1&mAP\\
			\hline
			CAN \cite{Liu2017End2End} &60.3&35.9&72.1&47.9\\
			DNS \cite{Zhang2016}&61.02&35.68&71.56&46.03\\
			Gated Reid \cite{Varior2016Gated}&65.88&39.55&76.04&48.45\\
			MR B-CNN \cite{Ustinova2015}&66.36&85.01&90.17&41.17\\
			Cross-GAN \cite{Zhang2018Crossing}*&72.15&-&94.3&48.24\\
			SOMAnet ~\cite{Barros2018}&73.87&47.89&81.29&56.98\\
			HydraPlus-Net \cite{liu2017hydraplus}&76.9&91.3&94.5&-\\
			Verif.Identif \cite{zheng2016discriminatively}&79.51&59.87&85.47&70.33\\
			MSCAN \cite{Li2017DeepContext} &80.31&57.53&86.79&66.70\\
			SVDNet \cite{Sun2017SVDNet}&82.3&62.1&-&-\\
			DeepTransfer \cite{Geng2016}*&83.7&65.5&89.6&73.80\\
			LSRO \cite{zheng2017unlabeled}&83.97&66.07&88.42&76.10\\
			JLML \cite{Li2017DeepJoint} &85.1&65.5&89.7&74.5\\
			KFM-ReID \cite{Shen2018End2End} &90.1&75.3&-&-\\
			\hline \hline	
			\textbf{SAG}&\textbf{90.17}&\textbf{73.87}&\textbf{92.76}&\textbf{80.15}\\
			\textbf{SAG+RR}&\textbf{92.04}&\textbf{89.28}&\textbf{94.60}&\textbf{85.32}\\					
			\hline
		\end{tabular}
	\end{table}
	
	\textbf{Evaluation on Market-1501} We compared our model with existing works in Market-1501 datasets and showed the superiority of our model. We achieved a $90.17\%$ rank-1 accuracy and $73.87\%$ mAP on single query setting. Our method outperforms JLML \cite{Li2017DeepJoint}(hard attention) by a factor of $5.07\%$ and slightly outperforms KFM-ReID \cite{Shen2018End2End} (Residual Self Attention) by a factor of $0.07\%$ on rank-1 accuracy achieving state-of-arts on attention-based CNN models for person re-id.
	\begin{table}
		\centering
		\caption{Comparison result with state-of-arts on CUHK03.}
		\label{tab:cuhk03results}
		\begin{tabular}{|l|c|c|c|c|}
			\hline
			\multirow{2}{*}{Methods}&\multicolumn{4}{c|}{CUHK03}\\
			\cline{2-5}
			&R1&R5&R10&mAP \\
			\hline \hline
			SI-CI ~\cite{Wang2016}&52.20&84.30&94.8&-\\
			DNS ~\cite{Zhang2016}	&54.7&80.1&88.30&-\\
			FisherNet ~\cite{Wu2016FisherNet}&63.23&89.95&92.73&44.11\\
			MR B-CNN ~\cite{Ustinova2015}&63.67&89.15&94.66&-\\
			Gated ReID ~\cite{Varior2016Gated}&68.1&88.1& 94.6&58.8\\
			SOMAnet ~\cite{Barros2018}&72.40&92.10&95.80&-\\
			SSM ~\cite{Bai2017}&72.7&92.4&96.1&-\\
			SVDNet ~\cite{Sun2017SVDNet}&81.8&95.2&97.2&84.8\\
			\hline \hline
			\textbf{SAG}&\textbf{82.46}&\textbf{96.44}&\textbf{98.42}&\textbf{88.64}\\
			\hline
		\end{tabular}
	\end{table}
	
	\textbf{Evaluation on CUHK03} On this dataset, we achieved an $82.46\%$ rank-1 accuracy and $88.64\%$ mAP respectively. We improve the baseline by a factor of $11.42\%$ on rank-1 accuracy and $8.18\%$ on mAP respectively. 
	\begin{table}
		\centering
		\caption{Comparison results of the state-of-arts methods on DukeMTMCReID.}
		\label{tab:dukeresults}
		\begin{tabular}{|l|c|c|c|c|}
			\hline
			\multirow{2}{*}{Methods} & \multicolumn{4}{c|}{DukeMTMCReID}\\
			\cline{2-5}
			&R1&R5&R10&mAP \\
			\hline \hline
			PUL \cite{Fan2017PUL}&36.5&52.6&57.9&21.5\\
			SPGAN \cite{Weijian2018Image2Image}&46.9&62.6&68.5&26.4\\
			LSRO ~\cite{zheng2017unlabeled}&67.68&-&-&47.13\\
			OIM ~\cite{xiaoli2017joint}&68.1&-&-&47.4\\
			TriNet ~\cite{Hermans2017Triplet}*&72.44&-&-&53.50\\
			SVDNet~\cite{Sun2017SVDNet}&76.7&86.4&89.9&56.8\\
			DCC \cite{Wu2018CoAttention}&\textcolor{blue}{\textbf{80.3}}&\textcolor{blue}{\textbf{92.0}}&\textcolor{blue}{\textbf{97.1}}&\textcolor{blue}{\textbf{59.2}}\\
			\hline \hline
			\textbf{SAG}&\textbf{79.94}&\textbf{89.68}&\textbf{92.15}&\textbf{60.88}\\
			\textbf{SAG+RR}&\textbf{85.28}&\textbf{91.07}&\textbf{93.76}&\textbf{81.05}\\
			\hline
		\end{tabular}
	\end{table}
	
	\textbf{Evaluation on DukeMTMC-ReID} On this dataset, we achieved competitive result with DCC \cite{Wu2018CoAttention} which Co-Attention model exceeds our Attention-Grid by a small factor of $0.36\%$ $(79.94-80.3)$.
	
\begin{figure}[h]
		\centering
		\includegraphics[width=\linewidth]{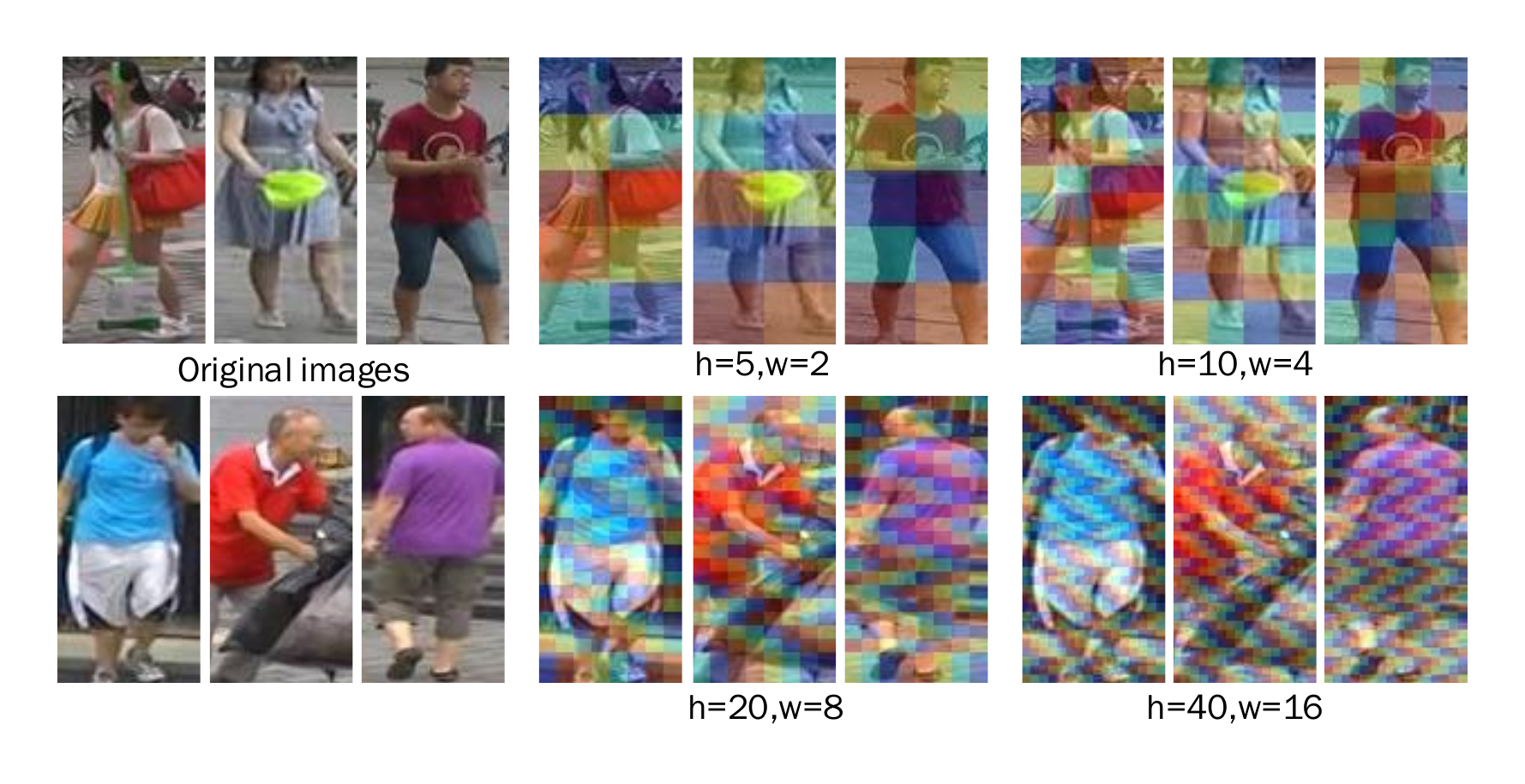}
		\caption{Visualization of the attention grid predicted by $D_1, D_2, D_3, D_4$ architectures. The first column shows he original images. The second and last columns show the attention grid learned with $h=5,10,20,40$ and $w=2,4,8,16$. High attentions are shown in red and yellow. The attention grid also predicts low values for backgrounds. Best viewed in color}
		\label{fig:multiregions}
	\end{figure}

	\textbf{Visualization of the Self Attention Grid}
	We visualize the attention grid at four different depths. Figure \ref{fig:multiregions} shows the visualization of our proposed attention mechanism. For example, with $(h=5,w=2)$, the results show how the SAG module can extract multi-parts and discriminative regions of the input images (e.g., backpack, legs, person's face, things in their hands, t-shirts). Also, it can be easily observed that our attention model successfully ignore the image's background.
	
	\subsection{Discussion} \label{discussion}
	An end to end CNN-based model which incorporates a self-attention mechanism at different levels has been proposed. The attention module can be plugged into any existing system and is fully differentiable. The parameters of the SAG module can be learned at the same time during training. In other words, the attention layers and the network are trained simultaneously using back-propagation. The main contribution of this paper is a deep attention grid that can focus on multiple regions of an image with high resolution and at the time preserve the internal information in the low resolution. 
	Previous approaches \cite{Li2017DeepJoint, Li2018Harmonious, Shen2018End2End, Wu2018CoAttention} for applying attention to re-identification mainly tend on finding a single attention region on the image for further processing; while it is true that a single regions can describe a person, we argue that paying attention to more than one region simultaneously can improve the person re-id. In general, we describe our proposed attention mechanism as modular, architecture independent, fast and simple. 
	
	During training, the gradient of our network can be decomposed into two additives terms with the first term propagating information directly through the first branch, without any attention information and the second term propagating information back to the attention zone units.
	This approach can be considered as a soft attention as the gradient is directly computed during training and the update of network parameters achieved through the use of the conventional Gradient Descent algorithm.
	
	We also observed classification accuracy improvement only when we add $L_2$ normalization to the $1^{st},2^{nd}$ and $3^{rd}$ residual units, but in $4^{th}$ layer, the classification accuracy drastically decrease. We therefore recommend using Equation \ref{eq:normalization} only on upper layers. 
	This is due to the non-sparseness propriety of the $L_2$ norm which positively affects the classification accuracy by leveraging high-level features but negatively disturb learned low-level features. 
	
	\section{Conclusion}
	This paper proposed Self Attention Grid (SAG) based-CNN model for person re-identification tasks. We proposed SAG modules to find the most informative regions of input images at different depth levels and combine them with the output feature maps of the same layer. Our proposed attention only uses the internal representation of the input vector at each step to update the attention response. We further performed an ablation study to demonstrate that the model generalizes well when applied to deep layers. Design choices for implementing the attention model (attention mode, weight sharing, output normalization and attention depth) has been proposed and compared using three popular datasets Market-1501 \cite{Zheng2015}, CUHK03 \cite{Li2014Pairing} and DukeMTMC-ReID \cite{zheng2017unlabeled}. 
	We successfully improve the accuracy of the baseline CNN model and outperform a vast range of state-of-art methods. In general, our attention grid mechanism can be adopted for any re-id task such as vehicle re-id. 
	\section{Acknowledgement}
	This work is supported by the Ministry of Science and Technology of Sichuan province (Grant No. 2017JY0073) and Fundamental Research Funds for the Central Universities
	in China (Grant No. ZYGX2016J083).
	
	{\small
		\bibliographystyle{ieee}
		\bibliography{egbib}

\begin{thebibliography}{10}\itemsep=-1pt

\bibitem{Bahdanau2015}
D.~Bahdanau, K.~Cho, and Y.~Bengio.
\newblock Neural machine translation by jointly learning to align and
  translate.
\newblock {\em ICLR}, Sept. 2015.

\bibitem{Bai2017}
S.~Bai, X.~Bai, and Q.~Tian.
\newblock Scalable person re-identification on supervised smoothed manifold.
\newblock In {\em 2017 IEEE Conference on CVPR}, pages 3356--3365, July 2017.

\bibitem{Barros2018}
I.~B. Barbosa, M.~Cristani, B.~Caputo, A.~Rognhaugen, and T.~Theoharis.
\newblock Looking beyond appearances: Synthetic training data for deep cnns in
  re-identification.
\newblock {\em Computer Vision and Image Understanding}, 167:50 -- 62, 2018.

\bibitem{Chen2016}
D.~Chen, Z.~Yuan, B.~Chen, and N.~Zheng.
\newblock Similarity learning with spatial constraints for person
  re-identification.
\newblock In {\em 2016 IEEE Conference on CVPR}, pages 1268--1277, June 2016.

\bibitem{Cheng2016}
D.~Cheng, Y.~Gong, S.~Zhou, J.~Wang, and N.~Zheng.
\newblock Person re-identification by multi-channel parts-based cnn with
  improved triplet loss function.
\newblock In {\em 2016 IEEE Conference on CVPR}, pages 1335--1344, June 2016.

\bibitem{Cheng2016LSTMMR}
J.~Cheng, L.~Dong, and M.~Lapata.
\newblock Long short-term memory-networks for machine reading.
\newblock In {\em EMNLP}, 2016.

\bibitem{Chorowski2015}
J.~Chorowski, D.~Bahdanau, D.~Serdyuk, K.~Cho, and Y.~Bengio.
\newblock Attention-based models for speech recognition.
\newblock In {\em NIPS}, pages 577--585, Cambridge, MA, USA, 2015. MIT Press.

\bibitem{Weijian2018Image2Image}
W.~Deng, L.~Zheng, Q.~Ye, G.~Kang, Y.~Yang, and J.~Jiao.
\newblock Image-image domain adaptation with preserved self-similarity and
  domain-dissimilarity for person re-identification.
\newblock In {\em CVPR}, 2018.

\bibitem{Fan2017PUL}
H.~{Fan}, L.~{Zheng}, and Y.~{Yang}.
\newblock {Unsupervised Person Re-identification: Clustering and Fine-tuning}.
\newblock {\em ArXiv e-prints}, May 2017.

\bibitem{Felzenszwalb2010}
P.~F. Felzenszwalb, R.~B. Girshick, D.~McAllester, and D.~Ramanan.
\newblock Object detection with discriminatively trained part-based models.
\newblock {\em IEEE Transactions on Pattern Analysis and Machine Intelligence},
  32(9):1627--1645, Sept 2010.

\bibitem{Geng2016}
M.~{Geng}, Y.~{Wang}, T.~{Xiang}, and Y.~{Tian}.
\newblock {Deep Transfer Learning for Person Re-identification}.
\newblock {\em ArXiv e-prints}, Nov. 2016.

\bibitem{Kaiming2015}
K.~He, X.~Zhang, S.~Ren, and J.~Sun.
\newblock Deep residual learning for image recognition.
\newblock In {\em 2016 CVPR}, pages 770--778, June 2016.

\bibitem{Hermann2015}
K.~M. Hermann, T.~Ko\v{c}isk\'{y}, E.~Grefenstette, L.~Espeholt, W.~Kay,
  M.~Suleyman, and P.~Blunsom.
\newblock Teaching machines to read and comprehend.
\newblock In {\em NIPS}, NIPS'15, pages 1693--1701, Cambridge, MA, USA, 2015.
  MIT Press.

\bibitem{Hermans2017Triplet}
A.~{Hermans}, L.~{Beyer}, and B.~{Leibe}.
\newblock {In Defense of the Triplet Loss for Person Re-Identification}.
\newblock {\em ArXiv e-prints}, Mar. 2017.

\bibitem{Goodfellow2014GAN}
G.~J. Ian, P.-A. Jean, M.~Mehdi, X.~Bing, S.~O. David, C.~Aaron, and B.~Yoshua.
\newblock Generative adversarial network.
\newblock In {\em NIPS}, 2014.

\bibitem{Jaderberg2015STN}
M.~{Jaderberg}, K.~{Simonyan}, A.~{Zisserman}, and K.~{Kavukcuoglu}.
\newblock Spatial transformer networks.
\newblock In {\em NIPS}, 2015.

\bibitem{Kostinger2012}
M.~Kostinger, M.~Hirzer, P.~Wohlhart, P.~M. Roth, and H.~Bischof.
\newblock Large scale metric learning from equivalence constraints.
\newblock In {\em 2012 IEEE CVPR}, pages 2288--2295, June 2012.

\bibitem{Li2017DeepContext}
D.~Li, X.~Chen, Z.~Zhang, and K.~Huang.
\newblock Learning deep context-aware features over body and latent parts for
  person re-identification.
\newblock In {\em 2017 IEEE Conference on CVPR}, pages 7398--7407, July 2017.

\bibitem{Li2014Pairing}
W.~Li, R.~Zhao, T.~Xiao, and X.~Wang.
\newblock Deepreid: Deep filter pairing neural network for person
  re-identification.
\newblock In {\em 2014 IEEE CVPR}, pages 152--159, June 2014.

\bibitem{Li2017DeepJoint}
W.~Li, X.~Zhu, and S.~Gong.
\newblock Person re-identification by deep joint learning of multi-loss
  classification.
\newblock In {\em Proceedings of the 26th International Joint Conference on
  Artificial Intelligence}, IJCAI'17, pages 2194--2200. AAAI Press, 2017.

\bibitem{Li2018Harmonious}
W.~Li, X.~Zhu, and S.~Gong.
\newblock Harmonious attention network for person re-identification.
\newblock In {\em The IEEE Conference on CVPR}, June 2018.

\bibitem{Liao2015}
S.~Liao, Y.~Hu, X.~Zhu, and S.~Z. Li.
\newblock Person re-identification by local maximal occurrence representation
  and metric learning.
\newblock In {\em 2015 IEEE CVPR}, pages 2197--2206, June 2015.

\bibitem{Liao2015a}
S.~Liao and S.~Z. Li.
\newblock Efficient psd constrained asymmetric metric learning for person
  re-identification.
\newblock In {\em 2015 IEEE International Conference on Computer Vision
  (ICCV)}, pages 3685--3693, Dec 2015.

\bibitem{Liu2017End2End}
H.~Liu, J.~Feng, M.~Qi, J.~Jiang, and S.~Yan.
\newblock End-to-end comparative attention networks for person
  re-identification.
\newblock {\em IEEE Transactions on Image Processing}, 26(7):3492--3506, July
  2017.

\bibitem{liu2017hydraplus}
X.~Liu, H.~Zhao, M.~Tian, L.~Sheng, J.~Shao, J.~Yan, and X.~Wang.
\newblock Hydraplus-net: Attentive deep features for pedestrian analysis.
\newblock In {\em Proceedings of the IEEE international conference on computer
  vision}, pages 350--359, 2017.

\bibitem{Mnih2014HardAttention}
V.~Mnih, N.~Heess, A.~Graves, and K.~Kavukcuoglu.
\newblock Recurrent models of visual attention.
\newblock In {\em NIPS}, NIPS'14, pages 2204--2212, Cambridge, MA, USA, 2014.
  MIT Press.

\bibitem{Paisitkriangkrai2015}
S.~Paisitkriangkrai, C.~Shen, and A.~van~den Hengel.
\newblock Learning to rank in person re-identification with metric ensembles.
\newblock In {\em 2015 IEEE Conference on Computer Vision and Pattern
  Recognition (CVPR)}, pages 1846--1855, June 2015.

\bibitem{Parikh2016Decomposable}
A.~Parikh, O.~T{\"a}ckstr{\"o}m, D.~Das, and J.~Uszkoreit.
\newblock A decomposable attention model for natural language inference.
\newblock In {\em Proceedings of the 2016 Conference on Empirical Methods in
  Natural Language Processing}, pages 2249--2255. Association for Computational
  Linguistics, 2016.

\bibitem{Parmar2018Transformer}
N.~{Parmar}, A.~{Vaswani}, J.~{Uszkoreit}, {\L}.~{Kaiser}, N.~{Shazeer},
  A.~{Ku}, and D.~{Tran}.
\newblock {Image Transformer}.
\newblock {\em ArXiv e-prints}, Feb. 2018.

\bibitem{Rahimpour2017Attention}
A.~Rahimpour, L.~Liu, A.~Taalimi, Y.~Song, and H.~Qi.
\newblock Person re-identification using visual attention.
\newblock In {\em 2017 IEEE International Conference on Image Processing
  (ICIP)}, pages 4242--4246, Sept 2017.

\bibitem{Ristani2016}
E.~Ristani, F.~Solera, R.~S. Zou, R.~Cucchiara, and C.~Tomasi.
\newblock Performance measures and a data set for multi-target, multi-camera
  tracking.
\newblock {\em Conference on Computer Vision workshop on Benchmarking
  Multi-Target Tracking}, abs/1609.01775, 2016.

\bibitem{Rodriguez2017AgeGender}
P.~Rodríguez, G.~Cucurull, J.~M. Gonfaus, F.~X. Roca, and J.~Gonzàlez.
\newblock Age and gender recognition in the wild with deep attention.
\newblock {\em Pattern Recognition}, 72:563 -- 571, 2017.

\bibitem{Shan2017Speech}
C.~{Shan}, J.~{Zhang}, Y.~{Wang}, and L.~{Xie}.
\newblock {Attention-Based End-to-End Speech Recognition on Voice Search}.
\newblock {\em ArXiv e-prints}, July 2017.

\bibitem{Shen2018End2End}
Y.~{Shen}, T.~{Xiao}, H.~{Li}, S.~{Yi}, and X.~{Wang}.
\newblock {End-to-End Deep Kronecker-Product Matching for Person
  Re-identification}.
\newblock In {\em The IEEE Conference on Computer Vision and Pattern
  Recognition (CVPR)}, June 2018.

\bibitem{Sun2017SVDNet}
Y.~Sun, L.~Zheng, W.~Deng, and S.~Wang.
\newblock Svdnet for pedestrian retrieval.
\newblock In {\em 2017 IEEE International Conference on Computer Vision
  (ICCV)}, pages 3820--3828, Oct 2017.

\bibitem{Sun2017RPP}
Y.~{Sun}, L.~{Zheng}, Y.~{Yang}, Q.~{Tian}, and S.~{Wang}.
\newblock {Beyond Part Models: Person Retrieval with Refined Part Pooling (and
  a Strong Convolutional Baseline)}.
\newblock {\em ArXiv e-prints}, Nov. 2017.

\bibitem{Szegedy2016}
C.~{Szegedy}, V.~{Vanhoucke}, S.~{Ioffe}, J.~{Shlens}, and Z.~{Wojna}.
\newblock Rethinking the inception architecture for computer vision.
\newblock In {\em 2017 IEEE Conference on Computer Vision and Pattern
  Recognition (CVPR)}, pages 2818--2826, July 2016.

\bibitem{Ustinova2015}
E.~{Ustinova}, Y.~{Ganin}, and V.~{Lempitsky}.
\newblock {Multiregion Bilinear Convolutional Neural Networks for Person
  Re-Identification}.
\newblock {\em ArXiv e-prints}, Dec. 2015.

\bibitem{Varior2016Gated}
R.~R. Varior, M.~Haloi, and G.~Wang.
\newblock Gated siamese convolutional neural network architecture for human
  re-identification.
\newblock In {\em ECCV}, 2016.

\bibitem{Vaswani2017Attention}
A.~{Vaswani}, N.~{Shazeer}, N.~{Parmar}, J.~{Uszkoreit}, L.~{Jones}, A.~N.
  {Gomez}, L.~{Kaiser}, and I.~{Polosukhin}.
\newblock {Attention Is All You Need}.
\newblock In {\em NIPS}. The Neural Information Processing Systems, June 2017.

\bibitem{Wang2017ResAttention}
F.~Wang, M.~Jiang, C.~Qian, S.~Yang, C.~Li, H.~Zhang, X.~Wang, and X.~Tang.
\newblock Residual attention network for image classification.
\newblock In {\em Computer Vision and Pattern Recognition ({CVPR})}, pages
  6450--6458, 07 2017.

\bibitem{Wang2016}
F.~Wang, W.~Zuo, L.~Lin, D.~Zhang, and L.~Zhang.
\newblock Joint learning of single-image and cross-image representations for
  person re-identification.
\newblock In {\em 2016 IEEE Conference on Computer Vision and Pattern
  Recognition (CVPR)}, pages 1288--1296, June 2016.

\bibitem{Wang2018}
J.~Wang, S.~Zhou, J.~Wang, and Q.~Hou.
\newblock Deep ranking model by large adaptive margin learning for person
  re-identification.
\newblock {\em Pattern Recognition}, 74:241 -- 252, 2018.

\bibitem{Xiaolong2018NonLocal}
X.~Wang, R.~Girshick, A.~Gupta, and K.~He.
\newblock Non-local neural networks.
\newblock {\em CVPR}, 2018.

\bibitem{Wu2016FisherNet}
L.~Wu, C.~Shen, and A.~Hengel.
\newblock Deep linear discriminant analysis on fisher networks: A hybrid
  architecture for person re-identification.
\newblock 65, 06 2016.

\bibitem{Wu2018CoAttention}
L.~{Wu}, Y.~{Wang}, J.~{Gao}, and D.~{Tao}.
\newblock {Deep Co-attention based Comparators For Relative Representation
  Learning in Person Re-identification}.
\newblock {\em ArXiv e-prints}, Apr. 2018.

\bibitem{xiaoli2017joint}
T.~Xiao, S.~Li, B.~Wang, L.~Lin, and X.~Wang.
\newblock Joint detection and identification feature learning for person
  search.
\newblock In {\em CVPR}, 2017.

\bibitem{Xiong2014}
F.~Xiong, M.~Gou, O.~Camps, and M.~Sznaier.
\newblock Person re-identification using kernel-based metric learning methods.
\newblock In {\em Computer Vision -- ECCV 2014}, pages 1--16, Cham, 2014.
  Springer International Publishing.

\bibitem{Xu2016AskAttendAnswer}
H.~Xu and K.~Saenko.
\newblock Ask, attend and answer: Exploring question-guided spatial attention
  for visual question answering.
\newblock In {\em Computer Vision -- ECCV 2016}, pages 451--466, Cham, 2016.
  Springer International Publishing.

\bibitem{Xu2015ShowAttendTell}
K.~Xu, J.~Ba, R.~Kiros, K.~Cho, A.~Courville, R.~Salakhudinov, R.~Zemel, and
  Y.~Bengio.
\newblock Show, attend and tell: Neural image caption generation with visual
  attention.
\newblock In {\em Proceedings of the 32nd International Conference on Machine
  Learning}, volume~37, pages 2048--2057, Lille, France, 07--09 Jul 2015. PMLR.

\bibitem{Yang2016StackedAttention}
Z.~Yang, X.~He, J.~Gao, L.~Deng, and A.~Smola.
\newblock Stacked attention networks for image question answering.
\newblock In {\em 2016 IEEE Conference on Computer Vision and Pattern
  Recognition (CVPR)}, pages 21--29, June 2016.

\bibitem{yu2017cross}
H.-X. Yu, A.~Wu, and W.-S. Zheng.
\newblock Cross-view asymmetric metric learning for unsupervised person
  re-identification.
\newblock In {\em Proceedings of the IEEE International Conference on Computer
  Vision}, 2017.

\bibitem{Zeyer2018Speech}
A.~{Zeyer}, K.~{Irie}, R.~{Schl{\"u}ter}, and H.~{Ney}.
\newblock {Improved training of end-to-end attention models for speech
  recognition}.
\newblock {\em ArXiv e-prints}, May 2018.

\bibitem{Zhang2018Crossing}
C.~{Zhang}, L.~{Wu}, and Y.~{Wang}.
\newblock {Crossing Generative Adversarial Networks for Cross-View Person
  Re-identification}.
\newblock {\em ArXiv e-prints}, Jan. 2018.

\bibitem{Zhang2016}
L.~Zhang, T.~Xiang, and S.~Gong.
\newblock Learning a discriminative null space for person re-identification.
\newblock In {\em 2016 IEEE CVPR}, pages 1239--1248, June 2016.

\bibitem{Zhang2016Picking}
X.~Zhang, H.~Xiong, W.~Zhou, W.~Lin, and Q.~Tian.
\newblock Picking deep filter responses for fine-grained image recognition.
\newblock In {\em 2016 IEEE Conference on Computer Vision and Pattern
  Recognition (CVPR)}, pages 1134--1142, June 2016.

\bibitem{Zhang2011}
Y.~Zhang and S.~Li.
\newblock Gabor-lbp based region covariance descriptor for person
  re-identification.
\newblock In {\em 2011 Sixth International Conference on Image and Graphics},
  pages 368--371, Aug 2011.

\bibitem{Zhao2017Diversified}
B.~Zhao, X.~Wu, J.~Feng, Q.~Peng, and S.~Yan.
\newblock Diversified visual attention networks for fine-grained object
  classification.
\newblock {\em IEEE Transactions on Multimedia}, 19(6):1245--1256, June 2017.

\bibitem{Zhao2017PartAligned}
L.~Zhao, X.~Li, Y.~Zhuang, and J.~Wang.
\newblock Deeply-learned part-aligned representations for person
  re-identification.
\newblock In {\em The IEEE International Conference on Computer Vision (ICCV)},
  pages 3219--3228, 2017.

\bibitem{Zheng2015}
L.~Zheng, L.~Shen, L.~Tian, S.~Wang, J.~Wang, and Q.~Tian.
\newblock Scalable person re-identification: A benchmark.
\newblock In {\em ICCV}, pages 1116--1124, Dec 2015.

\bibitem{zheng2016discriminatively}
Z.~Zheng, L.~Zheng, and Y.~Yang.
\newblock A discriminatively learned cnn embedding for person
  re-identification.
\newblock {\em ACM Transactions on Multimedia Computing Communications and
  Applications}, 2017.

\bibitem{zheng2017unlabeled}
Z.~Zheng, L.~Zheng, and Y.~Yang.
\newblock Unlabeled samples generated by gan improve the person
  re-identification baseline in vitro.
\newblock In {\em Proceedings of the IEEE International Conference on Computer
  Vision}, 2017.

\bibitem{Zhong2017reranking}
Z.~Zhong, L.~Zheng, D.~Cao, and S.~Li.
\newblock Re-ranking person re-identification with k-reciprocal encoding.
\newblock {\em Proceedings of the IEEE Conference on Computer Vision and
  Pattern Recognition}, 2017.

\bibitem{Zhong2017}
Z.~{Zhong}, L.~{Zheng}, G.~{Kang}, S.~{Li}, and Y.~{Yang}.
\newblock {Random Erasing Data Augmentation}.
\newblock {\em ArXiv e-prints}, Aug. 2017.

\bibitem{zhong2018camera}
Z.~Zhong, L.~Zheng, Z.~Zheng, S.~Li, and Y.~Yang.
\newblock Camera style adaptation for person re-identification.
\newblock In {\em CVPR}, 2018.

\end{thebibliography}
	}
	
\end{document}